%% file: 0main.tex
\documentclass[runningheads]{llncs}
\usepackage{graphicx}

\usepackage{tikz}
\usepackage{comment}
\usepackage{amsmath,amssymb} 
\usepackage{color}

\usepackage[width=122mm,left=12mm,paperwidth=146mm,height=193mm,top=12mm,paperheight=217mm]{geometry}

\usepackage{subfiles}
\usepackage{hyperref}
\usepackage{subcaption}

\hypersetup{
    colorlinks=true,
    linkcolor=blue,
    filecolor=magenta,      
    urlcolor=magenta,
    citecolor=blue,
}

\begin{document}
\pagestyle{headings}
\mainmatter
\def\ECCVSubNumber{2162}  

\title{Enhancing the Association in Multi-Object Tracking via Neighbor Graph} 

\author{Tianyi LIANG\inst{1} \and
	Long LAN\inst{1} \and
	Zhigang LUO\inst{1} }
\authorrunning{LIANG et al.}
%
\institute{National University of Defense Technology\\ \email{\{liantianyi,long.lan,zgluo\}@nudt.edu.cn} }

\titlerunning{Enhancing the Association in Multi-Object Tracking via Neighbor Graph}
%

\maketitle

\begin{abstract}
Most modern multi-object tracking (MOT) systems follow the tracking-by-detection paradigm. It first localizes the objects of interest, then extracting their individual appearance features to make data association. The individual features, however, are susceptible to the negative effects as occlusions, illumination variations and inaccurate detections, thus resulting in the mismatch in the association inference. In this work, we propose to handle this problem via making full use of the neighboring information. Our motivations derive from the observations that people tend to move in a group. As such, when an individual target's appearance is seriously changed, we can still identify it with the help of its neighbors. To this end, we first utilize the spatio-temporal relations produced by the tracking self to efficiently select suitable neighbors for the targets. Subsequently, we construct neighbor graph of the target and neighbors then employ the graph convolution networks (GCN) to learn the graph features. To the best of our knowledge, it is the first time to exploit neighbor cues via GCN in MOT. Finally, we test our approach on the MOT benchmarks and achieve state-of-the-art performance in online tracking.

\end{abstract}
\section{Introduction}
\subfile{1introduction}

\section{Related Work}
\subfile{2relatedwork}

\section{Approach}
\subfile{3approach}

\section{Experiments}
\subfile{4experiments}

\section{Conclusion}
\subfile{5conclusion}

\clearpage

\bibliographystyle{splncs04}
\bibliography{egbib}
\end{document}

%% file: 1introduction.tex
Multi-Object Tracking (MOT) aims to predict the trajectories of all target objects in video sequences. It has been a long-standing research topic in computer vision since many applications, such as video surveillance, autonomous driving and sport event analysis are built on it. In recent years, due to the advance of high-performance object detection, the trackers following tracking-by-detection paradigm made remarkable progress and dominate this community. Nonetheless, tracking multiple objects accurately in complex real-world scenes is still very challenging.

The basic pipeline of tracking-by-detection is first localizing objects of interest in each video frame and then associating them with certain metrics to form the trajectories. Under the online tracking protocol, this pipeline can be concisely defined as associating detection responses in current frame to existing trajectories. To this end, most recent state-of-the-art trackers adopt the re-identification (ReID) model to extract the individual appearance features (embeddings) of each detection and take them as cues for data association. Due to the benefits from the rapid development of deep re-identification technique, this kind of appearance feature based association seems to be robust in most cases. However, it should be not ignored that this scheme runs upon an important prerequisite, and that is the bounding box of detection should be accurate enough. Once inaccurate, distractions will be brought into the feature extraction, resulting in the error associations. And the image blurring caused by target or camera motions can also degenerate the quality of embeddings. Unfortunately, these two harmful cases are almost inevitable at present, even the tracker are equipped with the most advanced object detector and de-noise model. More importantly, in real-world scenes the frequent occlusions, illumination variations and cluttered backgrounds often dramatically change the appearance of targets, which makes the association based on individual appearance features difficult to make correct inferences.

To cope with the aforementioned problems, some works introduce more sophisticated ReID models to improve the appearance feature learning. For examples, \cite{Tang2017} and \cite{Babaee2018} employ human pose information and binary body mask respectively to highlight the foreground image and filter the background noises. These attempts are interesting but limited effective because they still only focus on the individual appearance feature learning. The individual appearance features of each detection response are very susceptible to the negative effects from occlusions, illumination variations and inaccurate detections, thus they are not powerful enough under complex tracking scenes. So how to overcome the inherent defects of individual features and make the association more robust? We  find similar problems also once troubled the research of recommender system. In specific, the recommender model cannot precisely predict the target user’s preferences when only use his individual features, as the individual features are not informative enough and susceptible to noises. To remedy this, the model first finds the similar users of the target, called neighbors, and then utilize the neighbors’ features to enrich the target user’s features. This strategy is termed as collaborative filtering. Here we argue that the philosophy of collaborative filtering can be extended to enhance the association in tracking-by-detection paradigm.

In this work, inspired by the philosophy of collaborative filtering, we propose to enhance the association through making full use of the neighboring information, rather than solely focusing on the individual features. Our main idea is concisely shown in Fig. 1: for the tracking of multiple pedestrians, although the target pedestrian’s appearance at current frame is seriously changed by occlusions, we still can associate it with the correct trajectory since the neighboring pedestrians provide important complementary information. Here we term the detection and trajectory waiting to be matched as the targets for convenience. This motivation also conform with the observation that people tend to walk in a group, and the entire group is relatively stable and consistent in a long term. As such, when a part of the group suffers from the occlusions, illumination variations or inaccurate bounding boxes, we can identify them with help of their neighbors in the group. In other words, the features learned from groups are more powerful than the individuals. The neighboring tracked objects become a kind of attribute of the target to support the association.  


Nonetheless, how to fuse the features of targets and neighbors is non-trivial. In addition, in the multi-object tracking task, a large number of intra-objects (e.g., pedestrians) often simultaneously appear while most of them are noises for the targets. Therefore, it is necessary to efficiently filter the noises and select the most suitable neighbors. To achieve the two goals, we first design a spatio-temporal relation based strategy to select neighbors for the targets respectively. This strategy only uses the spatio-temporal information generated by the tracking self thus is lightweight and efficient. Then, with the selected neighbors, we construct a pair of neighbor graph for the target detection and trajectory. In specific, the nodes of a neighbor graph are the target (detection or trajectory) and its neighbors, all of the neighbor nodes link with the target node. We employ the graph convolutional networks (GCN) \cite{kipf2017semi} to extract the graph features and use it to compute the affinity. 

\begin{figure*}
	\centering
	\includegraphics[width=4.7in]{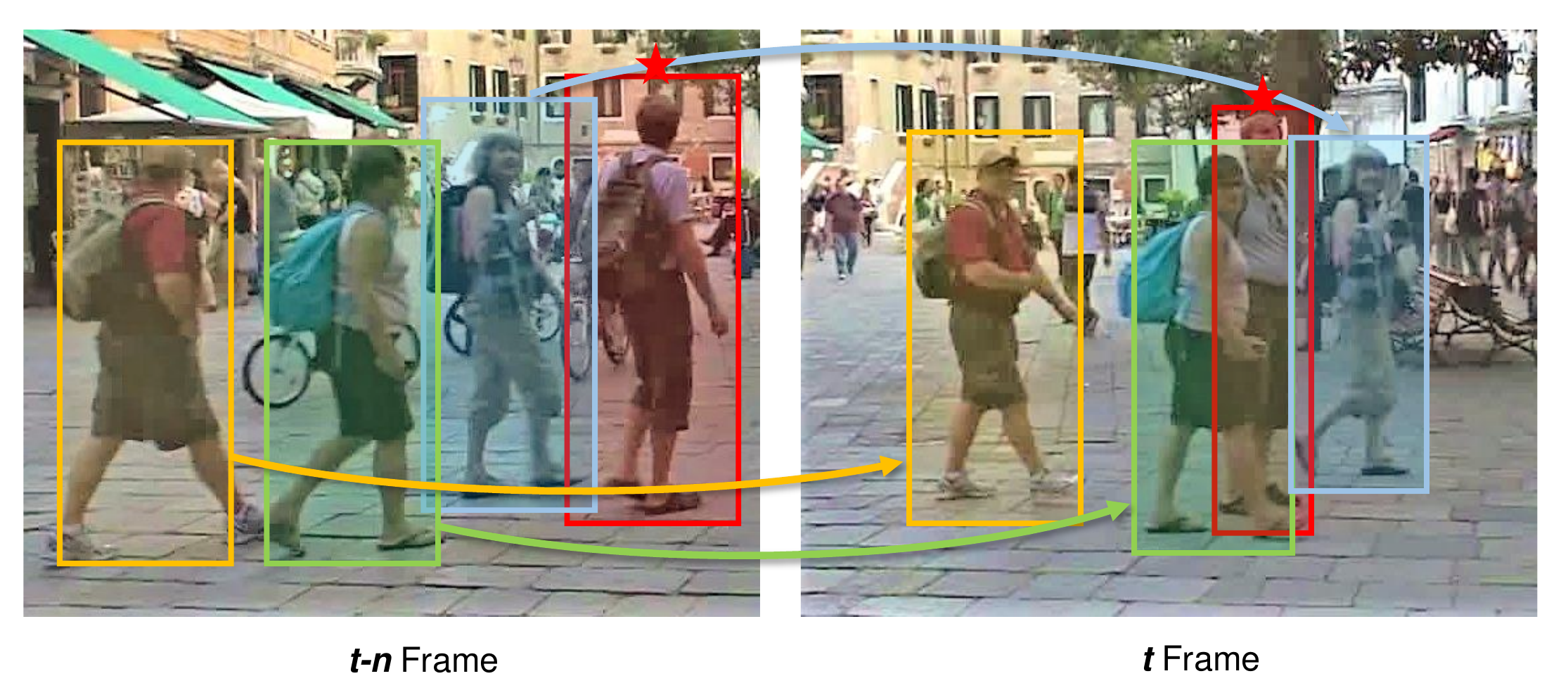}
	\caption{ Illustration of the neighbor-based association. The pedestrian with red bounding box and star is the ``target''. Although the target suffers serious occlusions at $t$ frame, we still can re-identify him with the help of neighbors jointly appear at frame $t-n$ frame and t frame. Best view in color.
	}
	\label{fig:main-idea}
\end{figure*}

We evaluate our approach on the most widely used MOT Challenge benchmark via the evaluation server. It achieves state-of-the-art performance on MOT16 \cite{milan2016mot16} and MOT17 \cite{milan2016mot16} datasets, following the online tracking protocol. In summary, our main contributions are: (1) We propose to enhance the association via jointly considering the target and its neighboring information. To the best of our knowledge, we are the first to exploit neighboring information in multi-object tracking. (2) We design a pragmatic method to select neighbors for the targets. It only uses the spatio-temporal cues generated by the tracking self, thus striking a good balance between accuracy and speed. Source codes of our approach will be released soon to support further research.

%% file: 2relatedwork.tex
As our contributions involve data association, re-identification and graph neural networks, we thus briefly review related works in these areas respectively.

\subsection{Data Association}
The tracking-by-detection framework consists of two components: an object detector to localize all the objects of interest, and a data association model to form the trajectories. In spite of the two parts are equivalently crucial for this task, most MOT works mainly concentrate on the data association because object detection is a separated research direction. Specifically, the association methods can be broadly classified in two categories: batch mode and online mode. The batch mode views tracking as a global optimization problem. It runs offline and utilizes cues from a long-time range of frames to output the final trajectories at once. A variety of global optimization algorithms such as graph segmentation \cite{Tang2016} and Marko random field \cite{lan2020semi-online} has been applied in this setting. In contrast, the online association process tracking as local optimization. It only focuses the association in two adjacent frames thus can be solved by the bipartite matching algorithms like Hungarian algorithm \cite{Bewley2016}. Compared with the batch mode association, online mode is more challenging since it cannot utilize the future frames to maintain the target identity in case of occlusions or detection missing. Although it is difficult, most state-of-the-art trackers dedicate to tracking online because this mode is much closer to the human-like ability. On the other hand, existing data association in both batch and online mode only consider the individual features of targets while ignoring its neighboring information. Our method fills this gap and improves the performance. Besides, in this work our method is implement and tested in online setting, but it can also be easily combined with the batch mode.

\subsection{ReID based Appearance Model}
Since the complex situations in real-world scenes, multiple cues including appearances \cite{Wojke2018}, motions \cite{Sadeghian2017} and interactions \cite{lan2016online} are jointly exploited to distinguish and re-identify targets. Among all of these, the appearance cues are most widely studied because the motions and interactions are hard to predict under long-term intra-object occlusions. In order to extract discriminative appearance features, most modern MOT trackers adopt the deep ReID model as the feature extractor. For examples, the DeepSORT tracker \cite{Wojke2018} employs a Resnet \cite{he2016deep} based ReID model to extract 128-dimension embeddings from detections and measures their affinities by the cosine distance. The ReID model is pretrained on a collection of large pedestrian ReID datasets, and this pretraining strategy now has been well-accepted in current MOT research. Compared with the SORT \cite{Bewley2016} which only uses the motion cues, DeepSORT considerably reduces the id-switches during tracking. The Siamese CNN architecture is used in \cite{kim2016similarity,leal-taixe2016learning}. Due to the contrastive training approach, the output appearance features are more discriminative. Works \cite{Tang2017,Babaee2018} try to further refine the feature embedding through reducing the noises from the background. \cite{Tang2017} trains the ReID model with human pose data to highlight the foreground of detection image patch. \cite{Babaee2018} uses Mask R-CNN \cite{he2017mask} to generate binary mask to filter the background image. Notably, although the aforementioned trackers can achieve comparative performance, their inference speed is usually very slow as the detecting and appearance embedding procedures are separated performed. To bring this gap, very recently some works unified the detection and embedding models, which is named Joint-Detection-Embedding (JDE), with respect to the Separate-Detection-Embedding (SDE) \cite{wang2019towards,Zhang2020A}. \cite{wang2019towards} first proposes to appends an embedding head on the heatmap of the YOLO detector \cite{redmon2018yolov3} then jointly train and test the entire model. As such, it can run at real-time speed without too much performance sacrifice. \cite{Zhang2020A} replaces the anchor-based detector \cite{redmon2018yolov3} with the anchor-free counterpart \cite{Zhou2019Objects}, such that alleviating the misalign of embedding feature. \cite{Zhang2020A} also uses a higher resolution heatmap to improve the quality embedding features. With these endeavors, \cite{Zhang2020A} achieves the best performance on MOT benchmark \cite{milan2016mot16} and runs at fast speed. Nonetheless, no matter SDE or JDE methods, existing works only focus the individual feature learning while ignoring the neighboring information, thus they are very susceptible to the harmful factors as occlusions and illumination variations. Instead, our solution incorporates the neighboring information to obtain more robust appearance features. Our model follows the JDE framework but also can be combined with the SDE.

\subsection{Graph Neural Networks}
Graph neural networks (GNN) \cite{kipf2017semi} is designed to work with the non-Euclidean data such as social relationships \cite{Hamilton2017Inductive}, molecular structures \cite{Kearnes2016Molecular} and knowledge graphs \cite{wang2018deep}. It has been applied in many fields to capture the complex interactions and relationships among objects. In computer vision, GNN has boost a series of tasks as semantic segmentation \cite{qi20173d}, action recognition \cite{yan2018spatial}, single object tracking \cite{gao2019graph} and person re-identification \cite{shen2018person}. For multi-object tracking, \cite{jiang2019graph} and \cite{ma2019deep} adopt GNN to perform the data association. They first extract targets’ appearance and motion features via CNN and LSTM respectively, then fusing them and make association inference through GNN. As GNN is differentiable, their entire frameworks thus can be trained in the end-to-end style. However, these works \cite{jiang2019graph,ma2019deep} still rely on the individual features and do not make full use of GNN to extracting more information from the tracked objects. In this work, we borrow some ideas from the work \cite{yan2019learning} and adopt graph convolution networks (GCN) \cite{kipf2017semi} to learn the features of our neighbor graph. GCN learns the relations in a graph with convolution operation, which facilitates the message pass and node updating. To the best of our knowledge, our work is the first to introduce GCN in MOT.

%% file: 3approach.tex
In this section we present our approach in detail. We first describe how to select suitable neighbors for the targets (section 3.1), then we introduce the methods of building the neighbor graph and learning graph features via GCN (section 3.2). Finally, we show the procedures of make data association using the neighbor graph features (section 3.3). The entire framework is depicted in Fig. 2.
\begin{figure*}
	\centering
	\includegraphics[width=4.7in]{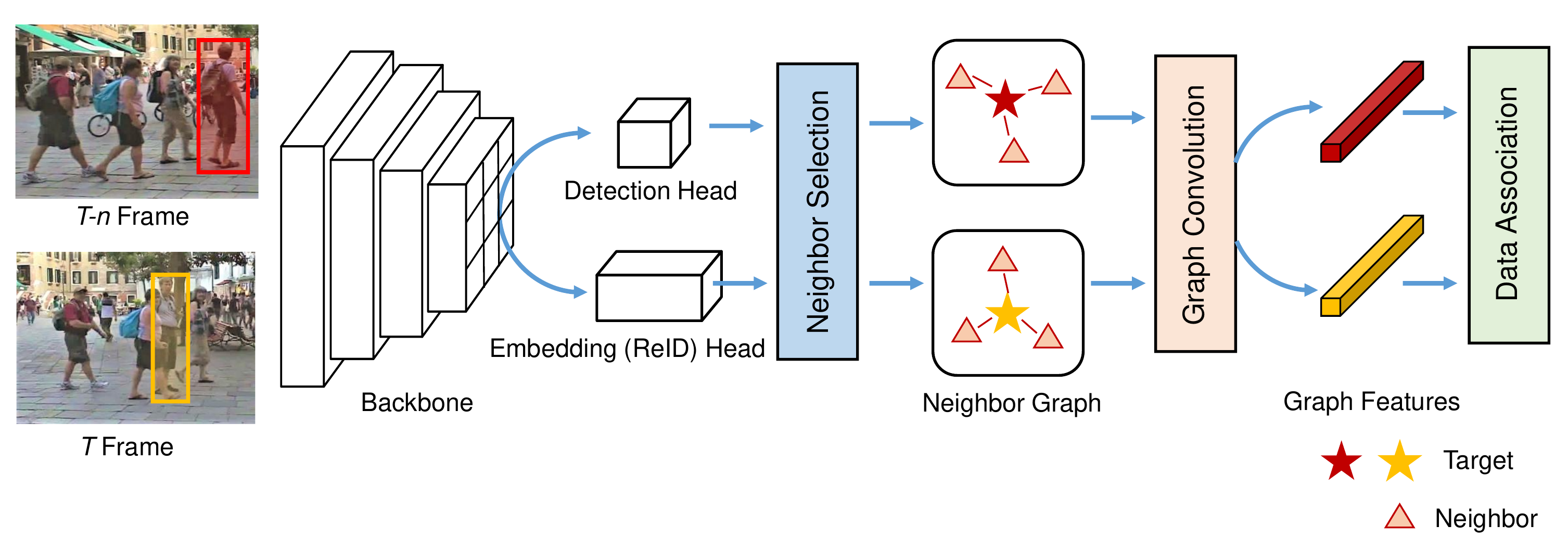}
	\caption{ The pipeline of our framework. We implement our framework in the Joint-Detection-Embedding style. The backbone first outputs the heatmap of the input image with down-sample 4. Then the detection and embedding heads localize the objects and extract their features. With these, we perform neighbor selection and build neighbor graph for the targets. Finally, we learn the graph features via GCN, and use them to make associations.
	}
	\label{fig:framwork}
\end{figure*}

\subsection{Spatio-Temporal Relation based Neighbor Selection}
Neighbor selection is crucial to our framework since many objects of interest usually simultaneously appear in a frame but most of them are irrelevant to the specific target pair (i.e., trajectory-detection pair). Therefore, it is necessary to filter the noises and pick up the most suitable ones as neighbors. Unlike some other fields, such as recommender system and social network analysis which can run complex neighbor-search procedures offline, multi-object tracking, especially in online mode, is a time-critical task thus its neighbor selection is expected to be time-efficient and needs no extra training data. To this end, we propose to efficiently make neighbor selection only using the spatio-temporal information produced by tracking self. 

In specific, at frame $I_t$, 
we first compute the affinity matrix according to individual appearance and motion cues then solve it with the Hungarian algorithm. This is the classic data association, and here we term it as initial association. The initial association outputs two sets: the matching set $M_t$ and unmatching set $U_t$. The prior set consists of matched trajectory-detection pairs while the latter contains the unmatched detections and trajectories. On this basis, we further refine it as: for any pair of trajectory-detection in $M_t$, it would be removed and put into $U_t$ if the corresponding affinity score is lower than the threshold $\tau^1$. This operation aims to make the initial association more conservative and reliable. Then, for each trajectory and detection in $U_t$, we select their suitable neighbors from the refined matching set $M_{t}^{'}$. Suppose we want to select neighbors for the trajectory $T_u$ and detection $d_u$ in the extended unmatching set $U_{t}^{'}$, we perform this process according to the temporal and spatial relations.

\textbf{Temporal relations:} neighbors of the targets should satisfy two temporal constraints: (1) the target (a detection or trajectory) and its neighbors should jointly appear in the same frame; (2) for any pair of targets, they should share the same neighbors at the time of matching, otherwise, they are inadequate to be matched using our neighbor graph. More formally, for an unmatching trajectory $T_u$ in $U_{t}^{'}$, we take its last active frame $I_{last}$ as the condition and search in $M_{t}^{'}$ to find the trajectories also used to active at frame $I_{last}$. The trajectories satisfying this temporal condition are called neighbor-candidates of $T_u$. With neighbor-candidates, then we take their new associated detections at frame $I_t$ as the neighbor-candidates for the unmatching detection $d_u$ in $U_{t}^{'}$. In other words, when we want to build a pair of neighbor graph for $T_u$ and $d_u$, their neighbors are all selected from the neighbor-candidates of $T_u$.

\textbf{Spatio relations:} observations from the pedestrian tracking scenario tell a fact that the spatial distances among the co-walkers are closer than the others. As such, we apply this rule to locate neighbors of the target and argue that it also can be extended to other tracking scenarios. Specifically, for the trajectory $T_u$ which last active at frame $I_{last}$, we compute its Euclidean distance to each item in $T_u$’s local neighbor-candidates, and select the top $K$ nearest candidates as the neighbors of $T_u$. Note that the distances are computed by the bounding boxes at frame $I_{last}$. The neighbors of $d_u$ are also selected in the same way, but the distances between and $d_u$ and its neighbor-candidates are computed by the bounding boxes at the current frame $I_t$.

\subsection{Learning Appearance Features from Neighbor Graph}

With the neighbors selected by the spatio-temporal relations, the following question is how to build the neighbor graph and effectively learn its appearance features. The optimal representation of the neighbor graph is expected to fully incorporate the neighboring information so that being discriminative enough and less affected by the harmful factors such as occlusions and inaccurate bounding boxes. To this end, we build the neighbor graph consisting of the target and corresponding neighbors, and employ GCN \cite{kipf2017semi} to learning the graph features. In order to facilitate the information propagation and feature updating among graph nodes, the target is placed in the center position and all the neighbors connect to it.

In specific, considering a neighbor graph $G$ consists of $N$ nodes and a set of edges. If the target in $G$ has $K$ neighbors, then $N=K+1$. For the $N$ nodes of $G$, they are assigned with the appearance feature vectors of the target and corresponding neighbors, i.e., $X \in R^{N \times d}$ where $d$ is the feature dimension of each node. As the target node may represent a detection or trajectory, thus the feature vectors input into $G$ are processed differently, depending on the type of the target. In particular, suppose we want to build a pair of neighbor graph $G_{det}$ and $G_{traj}$ for the detection $d_m$ at frame $I_t$ and the trajectory $T_n$ last active at frame $I_{t-n}$. For the $G_{det}$, the feature vectors associated with it are extracted from $d_m$ and its neighbors at frame $I_t$. But for the $G_{traj}$, the input feature vectors of $T_n$ and its neighbors are computed as function (1):
\begin{align}
f_t = \mu f_{t-1} + (1-\mu) \widetilde{f}
\end{align}
where $f_t$ is the smoothed feature of a trajectory at frame $I_t$, and $\widetilde{f}$ denotes the appearance feature of the associated detection in frame $I_t$. The momentum term $\mu$ is set to 0.9.

We use $A \in R^{N \times N}$ to denote the adjacent matrix of neighbor graph $G$. Let the target node as the first node in the $G$, then the adjacent matrix is:
\begin{align}
A_{i,j} = \begin{cases} 1, & \text{if} \; i=1 \; \text{or} j=1 \; \text{or} \; i=j ; \\ 0 & \text{otherwise},
\end{cases}
\end{align}
where $i,j \in \{1,...,N\}$. Let $\widetilde{A}$ denote the normalized adjacent matrix, the layer-wise propagations of GCN is computed as function (3):
\begin{align}
Z^{(l+1)} = \sigma(\widehat{A} Z^{(l)} W^{(l)})
\end{align}
where $Z^{(l)}$ is the activations of the $l$-th layer and $W^{(l)}$ is the learnable matrix. We use ReLU as the activation function $\sigma$. The network merges features of nodes and finally output a feature vector of 2048-dimension. At the training phase, the losses for backup propagations are computed by the cosine distances between predictions and labels.

\subsection{Association}

This round of data association is performed on the unmatched set in the initial association phase. For each trajectory and detection in the unmatched set, we build the neighbor graph for them and model their appearance features through GCN. With these graph features, we compute the affinity matrix and solve it using the Hungarian algorithm. The same as the post-process in the initial association, we filter the matching pairs which affinity scores are lower than the threshold $\tau^2$.

%% file: 4experiments.tex
\subsection{Datasets and Evaluation Metrics}
\textbf{Datasets.} As the prior JDE frameworks \cite{wang2019towards,Zhang2020A}, we train our entire model on a collection of object detection, person ReID and tracking datasets. In specific, we purely train the detection branch in our model on on the the ETH \cite{ess2008a} and the CityPerson \cite{zhang2017citypersons} datasets, while jointly train the ReID and detection branches on the datasets of CalTech \cite{dollar2009pedestrian}, MOT17 \cite{milan2016mot16}, CUHK-SYSU \cite{xiao2017joint} and PRW \cite{gheissari2006person}. For the testing, we evaluate our tracker on the MOT16 \cite{milan2016mot16} and MOT17 datasets which share the same 14 sequences of video (7 for training and 7 for testing). The differences of MOT16 and MOT17 datasets are the latter provides more kinds of public detections and finer ground truths.

\textbf{Evaluation metrics.} We adopt the CLEAR MOT Metrics \cite{bernardin2008evaluating} to evaluate our work. In specific, metrics used in our evaluations are multiple object tracking accuracy (MOTA), false positives (FP), false negatives (FN), identity switches (IDS), identification F1 score (IDF1), the number of mostly tracked targets (MT, $\textgreater$ 80\% recovered) and the number of mostly lost targets (ML, $\textless$ 20\% recovered). Among these metrics, MOTA summarizes FP, FN and IDS factors while seriously impressed by the first two. IDF1 can be complementary to MOTA since it emphasizes consistency (lower IDS). Both of them are most important for MOT trackers.

\subsection{Implementation Details}
We implement our tracker in the JDE framework with reference to the work \cite{Zhang2020A}. In particular, we use the modified DLA-34 network \cite{Zhou2019Objects} as our backbone. For an input image with the size of $H_{\text{image}} \times W_{\text{image}}$ , the backbone outputs a heatmap in shape of $C \times H_{\text{image}} /4 \times W_{\text{image}} /4$. The settings of the detection and embedding heads upon the heatmap layer are the same as \cite{Zhang2020A}. For the head of neighbor graph learning, it consists of 3 layers of GCN and is placed after the above two heads. This sub-network receives a neighbor graph containing 1 target and $K$ neighbors. When training and testing, if the number of neighbors is less than $K$, we copy the target to serve as neighbors. In the extreme case $K=0$, the neighbor graph is dropped. 

The DLA-34 backbone is initialized with the parameters pretrained on the COCO detection dataset \cite{lin2014microsoft}. The GCN model is pretrained as \cite{yan2019learning} on the person search dataset CUHK-SYSU \cite{xiao2017joint}. We train and finetune the entire model using the Adam optimizer for 30 epochs. The learning rate starts with 1e-4, then decays to 1e-5 and 1e-6 at 20 and 27 epochs. The input image is resized to 1088 $\times$ 608 and goes through a series of augmentation as scaling, rotation and jittering. The threshold $\tau^1$ and $\tau^2$ for associations are set to 0.85 and 0.95 respectively. The number of neighbors $K$ is set to 4.

\subsection{Comparison with the state-of-the-arts }

\setlength{\tabcolsep}{7pt}
\begin{table*}[!htb]
	\begin{center}
		\caption{Comparison with the state-of-the-arts under the ``public detection'' protocol. The symbol ``*'' means the trackor runs offline.}
		\label{table:sota}
		\begin{tabular}{l|l|l|lllll}
			\hline\noalign{\smallskip}
			Dataset & Tracker & Year & MOTA$\uparrow$ & IDF1$\uparrow$ & MT$\uparrow$ & ML$\downarrow$ & IDs$\downarrow$ \\
			\noalign{\smallskip}
			\hline
			\noalign{\smallskip}
			
			MOT16 & DASOT\cite{chu2020dasot} & 2020 & 46.1 & 49.4 & 14.6\% & 41.6\% & 802\\
			& MOTDT\cite{chen2018real} & 2018 & 47.6 & 50.9 & 15.2\% & 38.3\% & 792\\
			& LSST\cite{feng2019multi} & 2019 & 49.2 & 56.5 & 15.2\% & 38.3\% & 792\\
			& HDTR\cite{Babaee2018}\textsuperscript{*} & 2018 & 53.6 & 46.6 & 15.2\% & 38.3\% & 792 \\
			& Trackotr\cite{feng2019multi} & 2019 & 54.4 & 52.5 & 19\% & 36.9\% & 682 \\
			& Ours & 2020 & \textbf{57.7} & \textbf{62.6} & 18.8\% & \textbf{32.8\%} & 732 \\
			\noalign{\smallskip}
			\hline
			\noalign{\smallskip}
			MOT17 & DASOT\cite{chu2020dasot} & 2020 & 49.5 & 51.8 & 20.4\% & 34.6\% & 4142\\
			& MOTDT\cite{chen2018real} & 2018 & 50.9 & 52.7 & 17.5\% & 35.7\% & 2474\\
			& LSST\cite{feng2019multi} & 2019 & 54.7 & 62.9 & 20.4\% & 40.1\% & 3726\\
			& Trackotr\cite{feng2019multi} & 2019 & 53.5 & 52.3 & 19.5\% & 36.6\% & 2072 \\
			& TT\cite{zhang2020long}\textsuperscript{*} & 2020 & 54.9 & 63.1 & 24.4\% & 38.1\% & 1088 \\
			& Lif\_TsimInt\cite{hornakova2020lifted}\textsuperscript{*} & 2020 & 58.2 & 65.2 & 28.6\% & 33.6\% & 1022 \\
			& Ours & 2020 &{\bf 58.4} & 62.9 & 20.8\% & {\bf 31.3\%} & 2425 \\
			\hline
		\end{tabular}
	\end{center}
\end{table*}

\setlength{\tabcolsep}{7pt}
\begin{table}[!htbp]
	\begin{center}
		\caption{Comparison with JDE tracker \cite{Zhang2020A} under the ``private detection'' protocol. The results are evaluated on the motchallenge-devkit.}
		\label{table:oneshot}
		\begin{tabular}{c|c|cccccc}
			\hline\noalign{\smallskip}
			Dataset & Method & MOTA$\uparrow$ & FP$\downarrow$ & FN$\downarrow$& IDF1$\uparrow$ & IDs$\downarrow$\\
			\noalign{\smallskip}
			\hline
			\noalign{\smallskip}
			
			MOT17 $train$ & FairMOT \cite{Zhang2020A} & 76.4 & 18315 & 58827 & 73.0 & 2271\\
			& Ours & {\bf 76.6} & {\bf 18246} & {\bf 58578} & {\bf 73.6} & {\bf 2049}\\
			\hline
		\end{tabular}
	\end{center}
\end{table}
\setlength{\tabcolsep}{1.4pt}
We compare our method with the state-of-the-art trackers on the MOT benchmark. Results reported in Table 1 show that our tracker exceeds all other trackers on the MOT16 and MOT17 test sets. \cite{hornakova2020lifted} and \cite{zhang2020long} gain some advantages in terms of IDF1 and IDs, and that is because they run in offline mode thus can utilize the global frames of a video. Note that we evaluate our method under the public detection protocol, therefore we only keep the bounding boxes output by our model that are close to the public detections.

We also compare our method with the JDE tracker \cite{Zhang2020A}. For the sake of fairness, the settings, parameters, models and training strategies of our method and \cite{Zhang2020A} are identical. The only difference is our tracker equipped with the proporsed neighbor graph framework, and the results in Table 2 also demonstrate that the neighbor graph framework can significantly improve the data association procedures, empowering the tracker better ability to reduce id-switches.

%% file: 5conclusion.tex
The individual features of tracking targets are easily affected by the negatives as occlusions, pose variations and inaccurate detections, thus resulting in the mismatch of data association. In this work, we borrow some ideas from the collaborative filtering and propose to handle the aforementioned problem via exploiting the neighboring information. To this end, we first use the temporal and spatial cues from the tracking self to efficiently select suitable neighbors. Then, we build neighbor graph and employ GCN to learn graph embedding. Results on the MOT benchmark demonstrate our approach is effective. In the future, we consider further exploiting the neighboring cues to improve the object detection component in the Joint-Detection-Embedding framework.